\documentclass{article} %
\usepackage{colm2024_conference}

\usepackage{booktabs}
\usepackage{graphicx}
\usepackage{enumitem}
\usepackage{wrapfig}
\usepackage{algorithm}
\usepackage{algpseudocode}
\usepackage{multicol}

\usepackage{microtype}
\usepackage{amsmath}
\usepackage{colortbl}
\usepackage[utf8]{inputenc}
\definecolor{lightgray}{rgb}{0.9,0.9,0.9}
\usepackage{caption}
\usepackage{subcaption}
\usepackage{xcolor}
\usepackage{setspace}
\usepackage{url}
\usepackage{multirow}
\usepackage{colortbl}
\usepackage{tabularx}
\usepackage{blindtext}
\usepackage{pgfplots}
\pgfplotsset{compat=1.18} 
\usepackage{tikz}
\usetikzlibrary{er,positioning,bayesnet}
\usepackage{makecell}
\usepackage{tipa}
\usepackage{siunitx}
\usepackage{nicefrac}
\usepackage{tocloft}
\usepackage{listings}
\usepackage[raster,skins]{tcolorbox} %
\usepackage{xltabular}
\usepackage{adjustbox}
\usepackage{xurl}
\usepackage{setspace}

\usepackage{lipsum}
\usepackage{arydshln} 

\usepackage{amsmath,amsfonts,bm}

\def\eqref#1{equation~\ref{#1}}

\def\1{\bm{1}}

\DeclareMathAlphabet{\mathsfit}{\encodingdefault}{\sfdefault}{m}{sl}
\SetMathAlphabet{\mathsfit}{bold}{\encodingdefault}{\sfdefault}{bx}{n}

\title{UloRL:An Ultra-Long Output Reinforcement Learning Approach for Advancing Large Language Models' Reasoning Abilities}

\author{
\text{Dong Du}$^*$, \text{Shulin Liu}$^*$, \text{Tao Yang}$^*$, \text{Shaohua Chen}, \text{Yang Li} \\
Tencent Hunyuan Team \\
\texttt{\{dongdu,forestliu,rigorosyang,fafachen,youngyli\}@tencent.com} \\
}

\begin{document}

\maketitle

\begin{center}
    \textsuperscript{*}\textit{Contribute equally to this work.}
\end{center}

\begin{abstract}
Recent advances in large language models (LLMs) have highlighted the potential of reinforcement learning with verifiable rewards (RLVR) to enhance reasoning capabilities through extended output sequences. However, traditional RL frameworks face inefficiencies when handling ultra-long outputs due to long-tail sequence distributions and entropy collapse during training. To address these challenges, we propose an \textbf{U}ltra-\textbf{L}ong \textbf{O}utput \textbf{R}einforcement \textbf{L}earning (UloRL) approach for advancing large language models' reasoning abilities. Specifically, we divide ultra long output decoding into short segments, enabling efficient training by mitigating delays caused by long-tail samples. Additionally, we introduce dynamic masking of well-\textbf{M}astered \textbf{P}ositive \textbf{T}okens (MPTs) to prevent entropy collapse. Experimental results demonstrate the effectiveness of our approach. On the Qwen3-30B-A3B model, RL with segment rollout achieved 2.06x increase in training speed, while RL training with 128k-token outputs improves the model's performance on AIME2025 from 70.9\% to 85.1\% and on BeyondAIME from 50.7\% to 61.9\%, even surpassing Qwen3-235B-A22B with remarkable gains. These findings underscore the potential of our methods to advance the reasoning capabilities of LLMs with ultra-long sequence generation. We will release our code and model for further use by the community\footnote{\url{https://github.com/liushulinle/ULORL}}.

\end{abstract}
\begin{figure}[!h]    
\centering    
\includegraphics[width=0.99\textwidth]{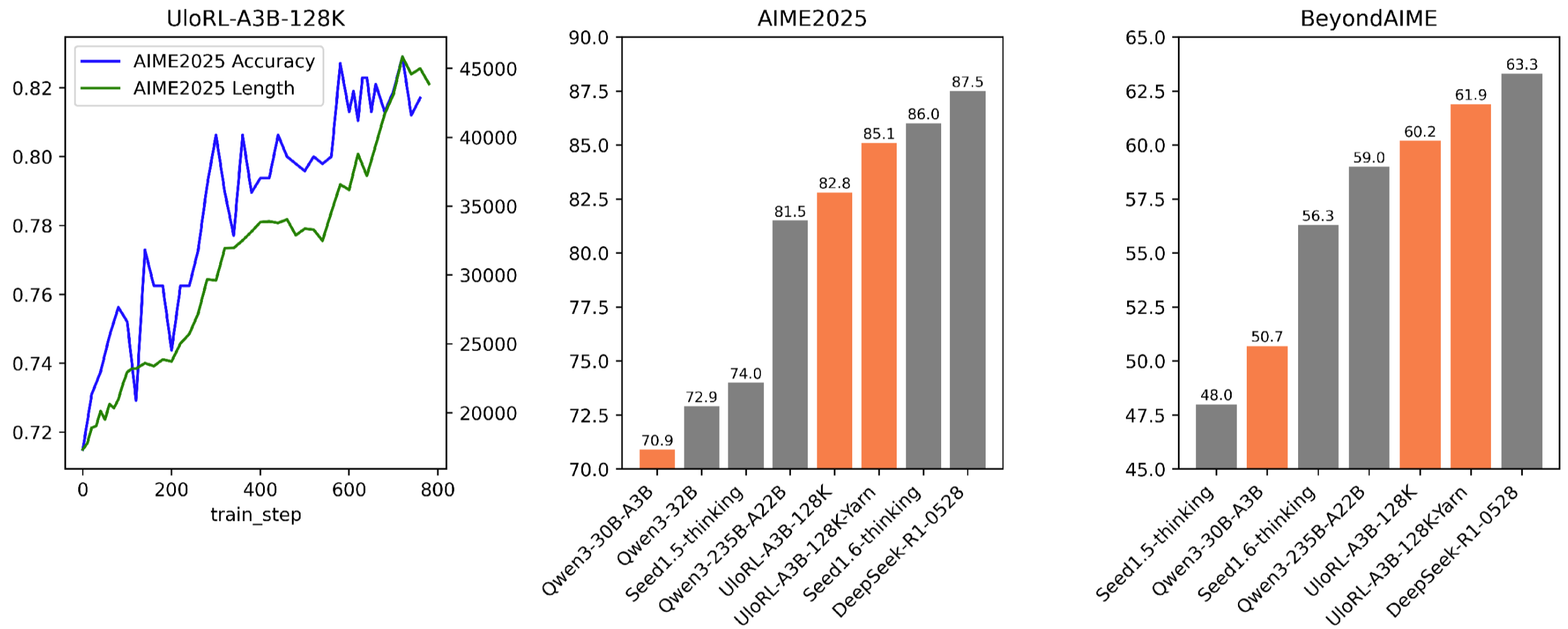}    
\end{figure}

\section{Introduction}
Recent advances in large language models (LLMs) have significantly enhanced their reasoning capabilities across challenging domains such as mathematics and programming. This progress has been driven by state-of-the-art models like OpenAI o1~\citep{jaech2024openai}, DeepSeek R1~\citep{guo2025deepseek}, and other models that employ sophisticated test-time scaling strategies. A key breakthrough in this evolution is the adoption of reinforcement learning with verifiable rewards (RLVR)\textbf{}. Unlike traditional reward shaping methods that focus on intermediate reasoning steps, this approach leverage rule-based verification systems to directly assess final answers, creating a powerful learning signal that guides the model toward generating correct and well-justified solutions through extended reasoning chains.

One of the key observations in recent advancements is that increasing the output length of models can significantly enhance their reasoning capabilities. However, traditional RL frameworks are not well-suited for such scenarios. In these frameworks, all samples in a batch must complete their decoding before training can proceed, leading to inefficiencies when dealing with long-tail distributions of sequence lengths. This inefficiency becomes particularly problematic when dealing with ultra-long outputs, such as outputs of up to 128k tokens, where a small fraction of long-tail samples can bottleneck the entire training process. 

K1.5~\citep{team2025kimi} proposed partial rollouts to address the aforementioned challenge. However, due to the lack of detailed descriptions of training strategies for segments from various models, as well as the absence of the setting of hyperparameters, it is challenging to reproduce their method. Similar with K1.5, we propose segment rollout which divides the decoding process into multiple stages. By decoding only a much shorter segment at each step, our method allows samples that have completed decoding to enter the experience pool for training immediately, while unfinished samples continue decoding in subsequent iterations. This approach not only accelerates training by avoiding unnecessary delays caused by long-tail samples but also ensures efficient utilization of computational resources. Furthermore, we introduce Segment-Aware Importance Sampling (SAIS) and Pesudo On-Policy Importance Sampling (POIS) to adapt the importance sampling mechanism to the segment rollout setting, ensuring accurate and stable training dynamics. We will release our code for further use by the community.

 Another critical challenge in RL training is the phenomenon of entropy collapse \citep{cheng2025reasoning,he2025skywork,yu2025dapo,zhu2025surprising}, where the model's diversity diminishes prematurely, leading to suboptimal performances. Existing research on addressing this issue can be broadly categorized into two approaches. The first approach involves directly incorporating an entropy loss term into the overall loss function, treating entropy as an additional optimization objective for the model ~\citep{guo2025deepseek,he2025skywork,wu2025confucius3,cheng2025reasoning}. However, since the goal of maintaining entropy is not fully aligned with the goal of improving reasoning abilities, this method may potentially hurt the model's performance ceiling. The second approach focuses on adjusting the samples or tokens involved in training ~\citep{yu2025dapo,zhu2025surprising,wang2025beyond}. For instance, DAPO~\citep{yu2025dapo} proposed increasing the clipping threshold to allow tokens with greater divergence from the current policy distribution to participate in training. However, this method is only effective in off-policy training, as on-policy training does not include a clipping mechanism. W-Reinforce~\citep{zhu2025surprising} proposed addressing the issue of entropy collapse by reducing the weight of positive samples duiring training. However, if the generation probabilities of certain important tokens within positive samples are inherently low, reducing the training weight in such cases may slow down the model's learning process and could even hurt the final performance.
 
In this work, we argue that the entropy collapse issue arises when the model overfits to well-\textbf{M}astered \textbf{P}ositive \textbf{T}okens (MPTs), i.e., tokens that the model already predicts with high confidence. To mitigate this, we introduce the \textbf{D}ynamic \textbf{M}asking of welll-\textbf{M}astered \textbf{P}ositive \textbf{T}okens (DMMPTs) strategy, which adaptively controls the training of such tokens based on the model's current entropy. Specifically, if the model's entropy falls below a predefined threshold, the MPTs are masked and excluded from the training process. Otherwise, all tokens are included in the training. The proposed DMMPTs neither introduces additional optimization objectives nor relies on importance sampling, thereby avoiding the limitations associated with the aforementioned approaches. Experiments on Qwen3-4B,Qwen3-8B and Qwen3-30B-A3B\citep{yang2025qwen3} illustrate that DMMPTs enables the model to maintain entropy stable during the training process.

Furthermore, we introduce a generative verifier model~\citep{zhang2024generative} to enhance the accuracy of reward computation in RL training. Unlike traditional rule-based methods~\citep{yu2025dapo,deepscaler2025}, which are prone to misjudgments in complex scenarios, our verifier model leverages generative capabilities to determine the equivalence of predicted and reference answers. Furthermore, to ensure the quality of the reward signal, we also emphasize the importance of data cleaning and transformation, including filtering noisy data, removing questions containing multiple sub-questions, standardizing problem formats and simplifying reference answer.

We conducted a series of experiments to validate the effectiveness of the proposed method. We performed RL training with an output length of 128k on the Qwen3-30B-A3B model. After training, the model's performance on AIME2025 improved from 70.9\% to 85.1\%, and on BeyondAIME\citep{seed2025seed1}, it improved from 50.7\% to 61.9\%, even surpassing the Qwen3-235B-A22B model (AIME2025: 81.5\%, BeyondAIME: 59.0\%) with remarkable gains.

\section{Preliminary}
\subsection{PPO}
PPO~\citep{schulman2017proximal} introduces a clipped surrogate objective for policy optimization. By constraining the policy updates within a proximal region of the previous policy using clip operations, PPO stabilizes training and improves sample efficiency. Specifically, PPO updates the policy parameters $\theta$ by maximizing the following objective:
\begin{equation}
\mathcal{J}_{\text{PPO}}(\theta) = \mathbb{E}_{(q,a)\sim\mathcal{D}, o_{\le t}  \sim \pi_{\theta_{\text{old}}}(\cdot|q)}
\Bigg[
\min\bigg(
\frac{\pi_{\theta}(o_t\mid q,o_{<t})}{\pi_{\theta_{\text{old}}}(o_t\mid q,o_{<t})}\hat{A}_t,
\,\text{clip}\Big(
\frac{\pi_{\theta}(o_t\mid q,o_{<t})}{\pi_{\theta_{\text{old}}}(o_t\mid q,o_{<t})},
1 - \varepsilon,
1 + \varepsilon
\Big)\hat{A}_t
\bigg)
\Bigg]
\end{equation}

where $(q,a)$ denotes a question-answer pair from the data distribution $\mathcal{D}$, $\varepsilon$ represents the clipping threshold that bounds policy updates, and $\hat{A}_t$ is the estimated advantage at step $t$. The advantage estimator $\hat{A}_t$ is computed using Generalized Advantage Estimation (GAE)~\citep{schulman2015high}:
\begin{equation}
\hat{A}_{t}^{\text{GAE}(\gamma,\lambda)} = \sum_{l=0}^{\infty} (\gamma\lambda)^{l} \delta_{t+l}
\end{equation}
with the temporal difference term $\delta_l$ given by:
\begin{equation}
\delta_l = R_{l} + \gamma V(s_{l+1}) - V(s_{l}), \quad  0 \leq \gamma,\lambda \leq 1
\end{equation}
where $R_l$ denotes the reward, $V$ represents the value function, $\gamma$ is the discount factor, and $\lambda$ controls bias-variance tradeoff in advantage estimation.

\subsection{GRPO}
GRPO~\citep{shao2024deepseekmath} presents a group-relative advantage estimation alternative  to PPO that eliminates dependency on value functions.  For any question-answer pair $(q,a)$, the behavioral policy $\pi_{\theta_{\text{old}}}$ generates a group of $G$ distinct responses $\{o_i\}_{i=1}^G$. The advantage for the $i$-th response $\hat{\mathcal{A}}_{i,t}$ is derived through group-level normalization:
\begin{equation}
\hat{\mathcal{A}}_{i,t} = \frac{r_i - \text{mean}(\{\mathcal{R}_i\}_{i=1}^G)}{\text{std}(\{\mathcal{R}_i\}_{i=1}^G)}
\end{equation}
GRPO also adopts a clipped surrogate function,  together with explicit KL regularization against a reference policy:
\begin{equation}
\begin{aligned}
\mathcal{J}_{\text{GRPO}}(\theta) = \mathbb{E}_{\substack{(q,a)\sim\mathcal{D}, \{o_i\}_{i=1}^G\sim\pi_{\theta_{\text{old}}}(\cdot|q)}}\\
\Biggl[ \frac{1}{G}\sum_{i=1}^G \frac{1}{|o_i|} \sum_{t=1}^{|o_i|} \biggl( \min\Bigl( & r_{i,t}(\theta) \hat{\mathcal{A}}_{i,t},  
 \text{clip}\bigl(r_{i,t}(\theta), 1-\varepsilon, 1+\varepsilon\bigr) \hat{\mathcal{A}}_{i,t} \Bigr) 
 - \beta D_{\text{KL}}\left(\pi_\theta \parallel \pi_{\text{ref}}\right) \biggr) \Biggr]
\end{aligned}
\end{equation}
where the importance ratio $r_{i,t}(\theta)$ measures policy update magnitude:
\begin{equation}
r_{i,t}(\theta) = \frac{\pi_{\theta}(o_{i,t} \mid q, o_{i,<t})}{\pi_{\theta_{\text{old}}}(o_{i,t} \mid q, o_{i,<t})}
\end{equation}

\subsection{DAPO}
DAPO~\citep{yu2025dapo} proposed a series of effective modifications based on GRPO for large scale RL training, including dynamic sampling, token-level gradient loss, clip higher, overlong reward shaping and removing KL divergence. The final objective is as follows:
\begin{equation}
\begin{aligned}
\mathcal{J}_{\text{DAPO}}(\theta) = \mathbb{E}_{\substack{(q,a)\sim\mathcal{D}, \{o_i\}_{i=1}^G\sim\pi_{\theta_{\text{old}}}(\cdot|q)}}\\
\Biggl[ \frac{1}{\sum_{i=1}^{G}|o_i|}\sum_{i=1}^G \sum_{t=1}^{|o_i|} \biggl( \min\Bigl( & r_{i,t}(\theta) \hat{\mathcal{A}}_{i,t},  
 \text{clip}\bigl(r_{i,t}(\theta), 1-\varepsilon_{low}, 1+\varepsilon_{high}\bigr) \hat{\mathcal{A}}_{i,t} \Bigr) 
 \biggr) \Biggr]\\
 &\text{s.t.} \quad 0 < \left|\{o_{i} \mid \text{is\_equivalent}(a, o_{i})\}\right| < G
\end{aligned}
\end{equation}

Following DAPO, we adopt dynamic sampling, token-level gradient loss and removing KL divergence in our approach.

\section{UloRL}
We propose an \textbf{U}ltra-\textbf{L}ong \textbf{O}utput \textbf{R}einforcement \textbf{L}earning (UloRL) approach for advancing large language models' reasoning abilities. In this section, we will introduce the key techniques associated with UloRL. Our implementation is built on the verl framework\footnote{\url{https://github.com/volcengine/verl}}~\citep{sheng2024hybridflow}.
\subsection{Segment Rollouts with Pseudo On-policy Importance Sampling}
Increasing the output length of models can enhance their reasoning capabilities~\citep{team2025kimi}. However, in scenarios involving ultra-long outputs, such as sequences with a length of 128k, the long-tail effect becomes a significant bottleneck. For example, within a batch, 80\% of the samples may have lengths within 64k, but all samples must wait for the longest 128k output to complete before next decoding iteration. This greatly reduces training efficiency and resource utilization.

To address this challenge, we divide the decoding of an ultra-long output into multiple stages. In each stage, only a segment of the sequence is decoded. Samples that complete decoding are immediately added to the experience replay buffer for training, while incomplete samples are carried over to the next iteration, where the results from the previous stage are concatenated and decoding continues. 

\begin{algorithm}[H]
\caption{RL Training with Segment Rollouts}
\label{alg:segmented_decoding}
\begin{algorithmic}[1]
\State \textbf{Initialize:} 
\State \hspace{1em} $\text{unfinished\_pool} \gets \{\}$ \Comment{Samples that unfinished decoding}
\State \hspace{1em} $\text{experience\_pool} \gets \{\}$ \Comment{Samples ready for training}
\State \hspace{1em} $\text{global\_max\_seq\_len} \gets 128K$ \Comment{Assume the global maximum decoding length is 128K}
\State \hspace{1em} $\text{max\_segment\_count} \gets 8$ \Comment{Assume the sequence is divided into 8 segments}
\State \hspace{1em} $\text{each\_segment\_length} \gets \text{global\_max\_seq\_len/max\_segment\_count}$
\For{$\text{step} \in \{1, 2, \dots, \text{total\_steps}\}$}
    \State \hspace{2em} $\text{batch} \gets \text{rollout}(\{\text{unfinished\_pool}, \text{prompts}\}, \text{max\_len}=\text{each\_segment\_length})$ \Comment{Rollout}
    \State \hspace{2em} $\text{unfinished\_pool} \gets \text{update\_unfinished\_pool}(\text{batch, unfinished\_pool})$
    \State \hspace{2em} $\text{experience\_pool} \gets \text{update\_experience\_pool}(\text{batch}, \text{experience\_pool})$
    \State \hspace{2em} $\text{update\_model}(\text{experience\_pool})$ \Comment{Update model}
\EndFor
\end{algorithmic}
\end{algorithm}

\subsubsection{Segment Rollouts}
Algorithm \ref{alg:segmented_decoding} illustrates the RL training process with segment rollouts. As illustrated in line 8, the input for each decoding step comes from two sources: (1) unfinished samples from the previous rollout step, and (2) new prompts from the RL dataset. The decoding process terminates under one of the following three conditions:
\begin{itemize}
    \item \textbf{End-of-sequence (EOS) token is encountered} In this case, the sample is considered complete and is added to the experience pool for training.
    \item \textbf{Segment reaches the maximum segment length but not the global maximum length} This indicates that the sample is not yet fully decoded and will be added to the unfinished pool for continuation in the next step.
    \item \textbf{Global maximum length is reached} In this case, the sequence is truncated and added to the experience pool for training.
\end{itemize}

Assuming the global maximum length is set to 128k, and the segment count is set to 8. Then the model only needs to decode 128k/8=16k at a time to perform an update. This avoids the inefficiency caused by waiting for a few ultra long samples to complete decoding, significantly improving training efficiency. To evaluate the impact of segment rollout on training efficiency, we conducted experiments on Qwen3-30B-A3B with 64k output and the results are illustrated in Table \ref{tab:speed}. From the table we observe that training with two segments and four segments can improve the training speed by 1.6x and 2.06x, respectively.

\begin{table}[!t]
    \centering
        \begin{tabular}{l|l|c}
            \toprule
            \textbf{segment count}  &\textbf{time cost per step} & \textbf{speed} \\
            \midrule
            1 & 1240s & 1.0x\\
            2 & 774s & 1.6x\\
            4 & 601s & 2.06x\\
            \bottomrule
        \end{tabular}
    \caption{The impact of segment count on training speed.}
    \label{tab:speed}
\end{table}  

\subsubsection{Training}
\begin{figure}[!t]
    \centering
    \includegraphics[width=0.9\textwidth]{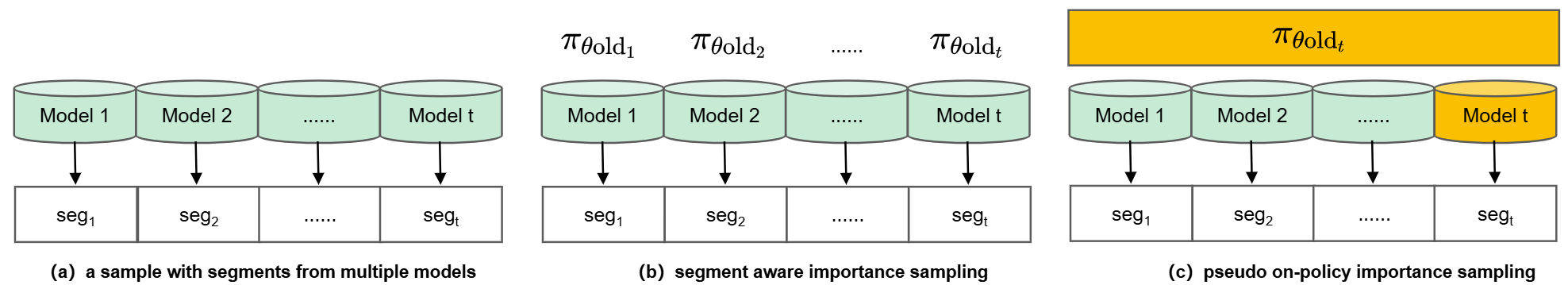}
    \caption{Illustration of a sample with segments from multiple models.}
    \label{fig:seg_rollout_1}
\end{figure}

In the original GRPO, each sample in the experience pool is generated by a single model. However, under the segment rollout setting, as illustrated in Figure \ref{fig:seg_rollout_1}(a), a single sample may consist of segments generated by multiple models. Consequently, the term $\pi_{\theta_{\text{old}}}$ in Equation 6 needs to be adjusted. To address this, we propose two methods for computing importance sampling value under segment rollout setting.

\textbf{Segment Aware Importance Sampling (SAIS)}
As illustrated in Figure \ref{fig:seg_rollout_1} (b), different segments are generated by different models, and therefore their corresponding $\pi_{\theta_{\text{old}}}$ vary. We denote the segment generated by the model at time $t$ as $\text{seg}_t$, then a sample $s$ can be represented as $s=[\text{seg}_1; \text{seg}_2; \dots;\text{seg}_t]$. For the $i$-th token in $s$, the importance sampling value can be computed using Equation \ref{eq:sais}, where $f(i)$ is the function to map the $i$-th token to its segment id.

\begin{equation}
r_{i,t}(\theta) = \frac{\pi_{\theta}(o_{i,t} \mid q, o_{i,<t})}{\pi_{\theta_{\text{old}_{f(i)}}}(o_{i,t} \mid q, o_{i,<|s_i|})}
\label{eq:sais}
\end{equation}

\textbf{Pseudo On-policy Importance Sampling (POIS)}
Recent work \citep{he2025skywork,hao2025policy} demonstrated that on-policy training exhibits more stable entropy and better performance than off-policy training. This is primarily because, in off-policy training, tokens that deviate significantly from the current policy are clipped by the clipping operation in Equation 7, which reduces the diversity of the model. In contrast, in on-policy training, all tokens are generated by the current model, therefore $\pi_\theta = \pi_{\theta_{\text{old}}}$. As a result, the importance sampling weight for all tokens is equal to 1, ensuring that no tokens are clipped. This allows the model to observe more diverse data during training. To leverage this advantage of on-policy training, we modify the importance sampling item to enable on-policy training.

As shown in Figure \ref{fig:seg_rollout_1} (c), at time step $t$, the last segment $\text{seg}_t$ is on-policy data, while segments generated from time steps 1 to $t-1$ are off-policy data. To achieve on-policy training, we simply replace the $\pi_{\theta_\text{old}}$ of all time steps with the $\pi_{\theta_{\text{old}_t}}$. Under this modification, the importance sampling weight for all tokens becomes 1. In this approach, samples with only one segment are true on-policy samples. For samples with more than one segment, the last segment is true on-policy, while other segments are pseudo on-policy. 

\begin{figure}[!t]
    \centering
    \includegraphics[width=0.8\textwidth]{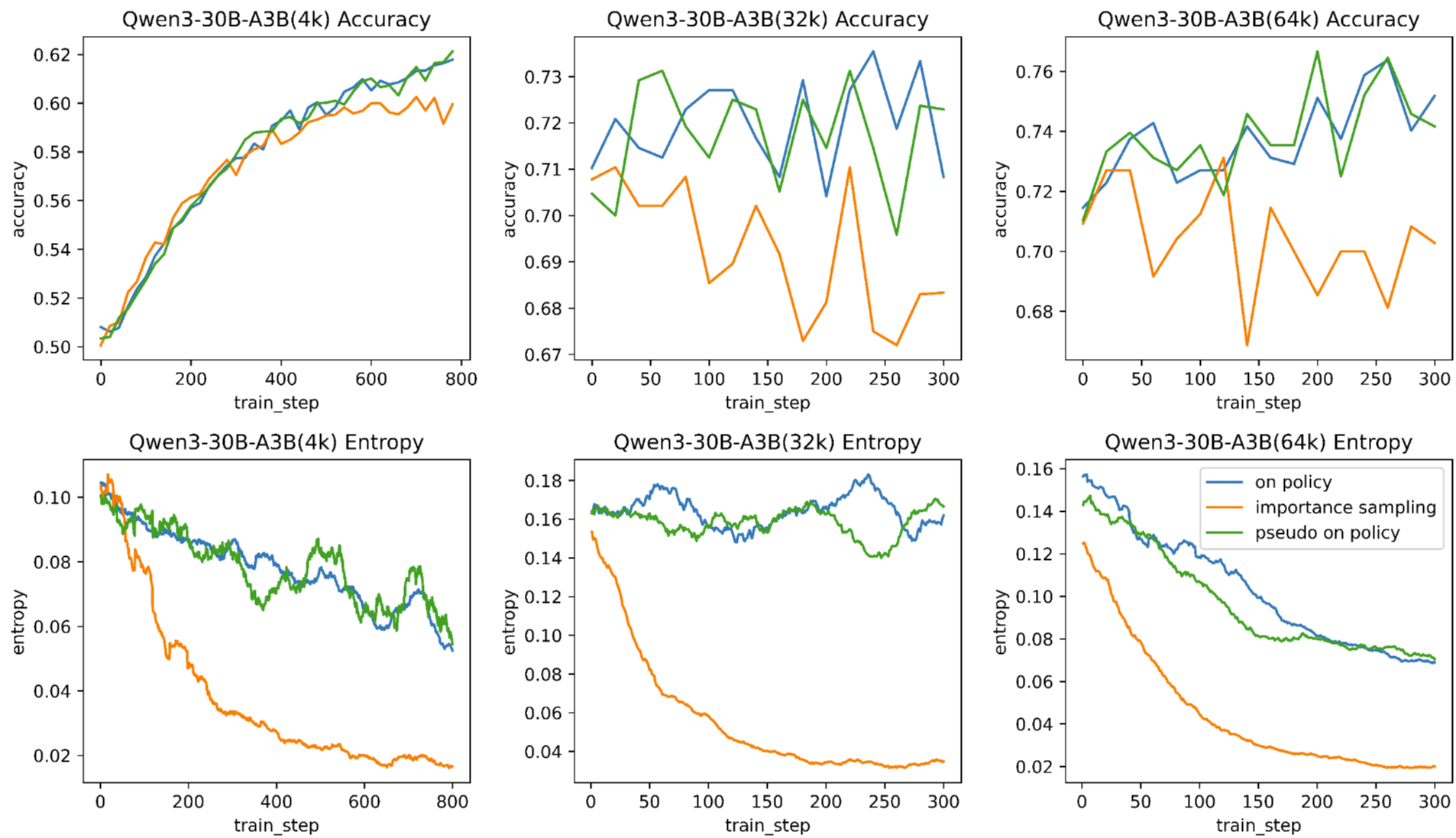}
    \caption{Training Dynamics of Different Importance Sampling Approaches.}
    \label{fig:is_compare}
\end{figure}

\textbf{Experimental Results}
To evaluate the effectiveness of the aforementioned methods, we conducted experiments on the Qwen3-30B-A3B model. The experimental settings are as follows:
\begin{itemize}
    \item \textbf{TOIS} True On-Policy Importance Sampling, with segment count $ = 1$
    \item \textbf{SAIS} Segment-Aware Importance Sampling, with segment count $ = 4$
    \item \textbf{POIS} Pseudo On-Policy Importance Sampling, with segment count $ = 4$
\end{itemize}
Figure \ref{fig:is_compare} illustrates the dynamics of entropy and accuracy for output lengths of 4k, 32k and 64k. Under the 4k output setting, it is surprising to observe that the entropy and evaluation curves of POIS and TOIS nearly overlap, and both outperform SAIS. Furthermore, the POIS also outperform SAIS with both 32k and 64k output. The effectiveness of POIS can potentially be attributed to the fact that the last segment of each sample is true on-policy data, which may mitigate the negative impact of training on pseudo on-policy data to some extent. Moreover, we also applied the Clip-higher strategy with $\epsilon_{high}=0.28$ as suggested in \cite{yu2025dapo} on SAIS. However, we observed entropy explosion, a phenomenon consistent with the findings of \cite{he2025skywork}.  \textbf{Based on these observations, we adopt POIS for subsequent experiments}.

\subsection{Avoiding Entropy Collapse: Do Not Train Well-mastered Positive Tokens}
\label{sec:mpts}
\subsubsection{Well-mastered Positive Tokens Result in Entropy Collapse}
\cite{zhu2025surprising} pointed out that training on positive samples is the primary cause of entropy reduction. We argue that the true reason is the overtraining of tokens that the model has already mastered within positive samples. Here, positive samples refer to those with a reward of 1, and "already mastered tokens" are defined as tokens for which the model's predicted probability exceeds a high threshold $\tau$. We refer to such tokens as well-\textbf{M}astered \textbf{P}ositive \textbf{T}okens (MPTs).
\begin{equation}
    \text{MPTs}=\bigcup\limits_{k=1}^{N} \bigcup\limits_{i=1}^{L_k}  \{t_i\in s_k\}, \text{where} \ \ p(t_i) \geq \tau, r(s_k)=1
    \label{EQ:MPTS}
\end{equation}

\begin{figure}[!h]
    \centering
    \includegraphics[width=0.8\textwidth]{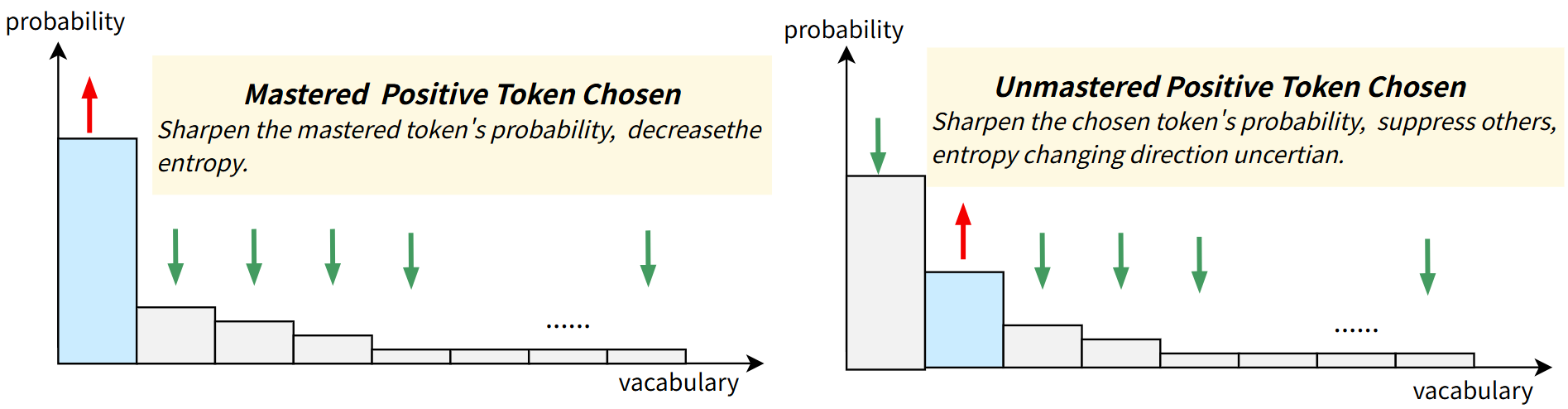}
    \caption{Entropy changing direction of updating MPTs (left) and non-MPTs (right), where blue block denotes the chosen token.}
    \label{fig:mpts}
\end{figure}

As shown in the left part of Figure \ref{fig:mpts}, updating the MPTs further increases its predicted probability. This, in turn, sharpens the distribution, making it more concentrated around the chosen token. As a result, the entropy of the model decreases. In contrast, as illustrated in the right part of Figure \ref{fig:mpts}, updating non-MPTs does not necessarily lead to a decrease in entropy.

We conducted experiments to validate the above hypothesis. The experimental setup is as follows:
\begin{itemize}
    \item \textbf{Baseline} All tokens are included in the training process.
    \item \textbf{Masking MPTs}: Only tokens excludes MPTs are included in the training process, where the threshold $\tau$ in Equation \ref{EQ:MPTS} is set to $0.99$. 
\end{itemize}

\begin{figure}[!t]
    \centering
    \includegraphics[width=0.7\textwidth]{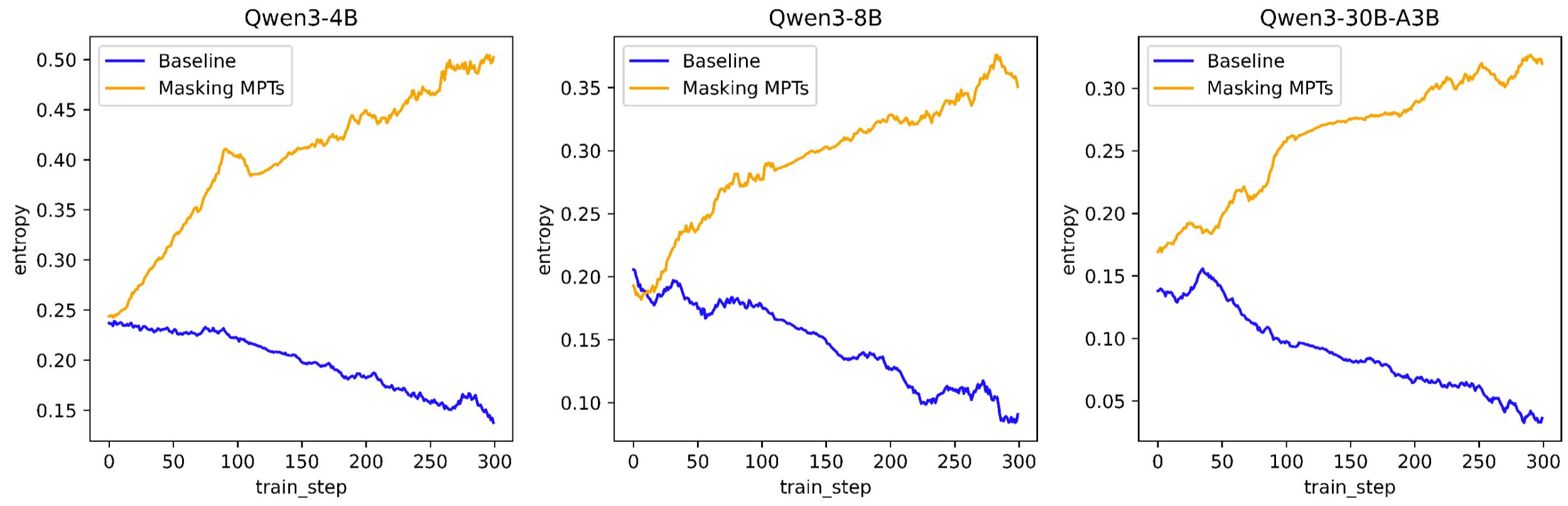}
    \caption{Training dynamics of RL with masking MPTs.}
    \label{fig:mpts_2}
\end{figure}

The experiments were performed on the Qwen3-4B, Qwen3-8B, and Qwen3-30B-A3B. The output length is set to 128k, which is divided into 8 segments. Figure \ref{fig:mpts_2} illustrates the entropy dynamics. As shown in the results, for all three models, the entropy of the baseline gradually decreases as training progresses. However, when MPTs are excluded from the training process, the entropy of the model increases over time. This observation supports our hypothesis that the overtraining of MPTs is a key factor contributing to the reduction in entropy. By excluding MPTs, the model maintains a more diverse output distribution.

\subsubsection{Dynamic Mask Well-mastered Positive Tokens}
As shown in Figure \ref{fig:mpts_2}, simply excluding MPTs from training leads to a continuous increase in entropy, which is also detrimental to the stability of training. Ideally, the model's entropy during training should be maintained around an appropriate entropy to ensure stable and effective learning. To achieve this, we propose a method called \textbf{D}ynamic \textbf{M}asking of MPTs (DMMPTs). Specifically, we introduce a new hyperparameter $\sigma$ to represent the target entropy. During training, MPTs are masked only when the current entropy falls below the target entropy $\sigma$: 

\begin{equation}
\begin{aligned}
\mathcal{J}(\theta) = &\mathbb{E}_{(q, a) \sim \mathcal{D},\ \{o_{i}\}_{i=1}^{G} \sim \pi_{\text{old}}}(\cdot \mid q) \\
&\left[\frac{1}{\sum_{i=1}^{G}|o_{i}|} \sum_{i=1}^{G} \sum_{t=1}^{|o_{i}|}[1-\mathbb{I}_\text{msk}] \min \left(r_{i, t}(\theta) \hat{A}_{i, t}, \operatorname{clip}\left(r_{i, t}(\theta), 1-\varepsilon, 1+\varepsilon\right) \hat{A}_{i, t}\right)\right] \\
\\
&\mathbb{I}_\text{msk}^{(i,t)}=
\begin{cases}
1 & \bar{\text{H}}_i < \sigma \ \text{and}\ o_i^t \in \text{MPTs} \\
0 & \text{otherwise}
\end{cases}
\\
&\bar{\text{H}}_i=\frac{1}{|o_i|}\sum_{t=1}^{|o_i|}\sum_{j=1}^{|\mathcal V|} p_t^{j}\log p_t^{j}
\end{aligned}
\end{equation}
This dynamic adjustment ensures that the model maintains a balanced entropy level, avoiding both excessive sharpness and excessive randomness in the output distribution. Note that, since MPTs have already been well-mastered by the model, it is expected that our approach will not have any negative impact on the model's performance.  

\begin{figure}[!t]
    \centering
    \includegraphics[width=0.7\textwidth]{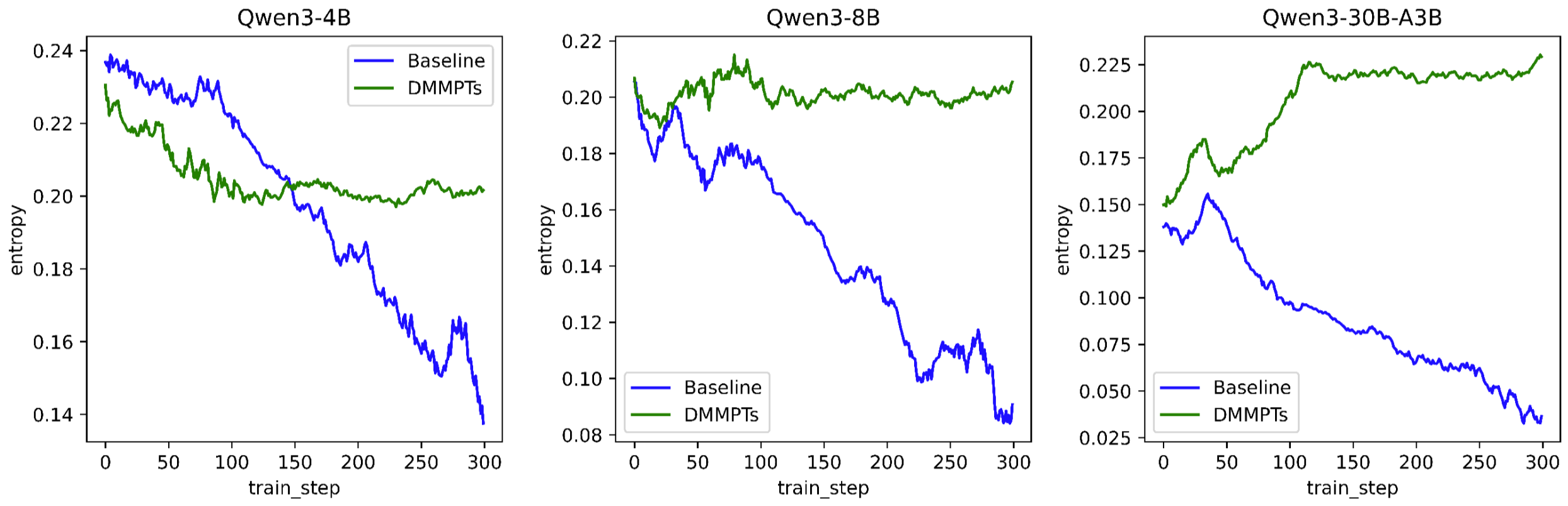}
    \caption{The entropy dynamics of DMMPTs.}
    \label{fig:mpts_3}
\end{figure}

To verify whether the proposed method achieves the desired objectives, we conducted experiments on the Qwen3-4B, Qwen3-8B, and Qwen3-30B-A3B. The experimental results are presented in Figure \ref{fig:mpts_3}. From the results, it can be observed that after incorporating the DMMPTs, the entropy of all three models, regardless of their size, remains stable around the predefined target range. This demonstrates the effectiveness of the proposed method in maintaining a balanced entropy level during training, thereby ensuring stable and robust learning.

\subsection{Generative Verifier Model}
Reward models based on outcomes have been proven to be highly effective for reinforcement learning (RL) in reasoning tasks~\citep{guo2025deepseek,yu2025dapo}. Following DAPO, we directly use the final accuracy of a verifiable task as the outcome reward. The reward is computed using the following rule:
\[
R(\widehat{y},y)=
\begin{cases} 
1, & \text{is\_equivalent}(\widehat{y},y) \\
0, & \text{otherwise.}
\end{cases}
\]
Unlike DAPO, our reward values are designed to be \(\{0, 1\}\) instead of \(\{-1, 1\}\). The advantage of this design is that the average reward across the dataset directly corresponds to its accuracy. Furthermore, under the GRPO framework, when both positive and negative samples exist within a group, the advantage for samples with a reward of 1 is always greater than 0, resulting in a positive gradient direction. Conversely, the advantage for samples with a reward of 0 is always less than 0, leading to a negative gradient direction. This behavior aligns well with the optimization objective.

Additionally, determining whether two answers \((\hat{y}, y)\) are equivalent is not a trivial task, as rule-based methods are prone to misjudgments. For example, pairs such as \((\text{27cm}, \text{0.27m})\) or \((\text{1/2}, \text{one half})\) can easily be misclassified as non-equivalent. To address this issue, we trained a generative model to evaluate whether two given answers are semantically equivalent. This approach ensures a more robust and accurate equivalence judgment.

\subsection{Data Cleaning and Transformation}
To ensure the accuracy of the rewards generated by the Verifier Model, we applied a series of preprocessing steps to the RL training data, addressing both the question and reference answer dimensions:

\textbf{Question Dimension}
\begin{itemize}
    \item \textbf{Deleting Problems of Multiple Sub-questions} We removed instances of multiple sub-questions within a single problem to avoid pseudo-negative reward caused by incomplete summaries of answers to sub-questions.
    \item \textbf{Converting Special Questions to Short-answer Format} We convert multiple-choice, proof-based and true/false questions to short-answer format to prevents the model from simply guessing the correct answer without understanding the problem.
    \item \textbf{Deleting Overly Simple Questions} To enhance the efficiency of reinforcement learning (RL) training, we utilized the Qwen3-30B-A3B model to perform inference on all data 8 times. Questions that were answered correctly in all 8 attempts were deemed overly simple and subsequently removed from the training dataset. 
\end{itemize}

\textbf{Answer Dimension}
\begin{itemize}
    \item \textbf{Extracting Short Answers For Reference Answers} This reduces the complexity of the verifier model's judgment task, improving its accuracy.

\item \textbf{Deleting Questions With Excessively Long Reference Answers} For example, problems with matrix-based answers were excluded to avoid unnecessary complexity for the verifier model.

\item \textbf{Deleting Questions With Incorrect Reference Answers} To identify such cases, we used multiple SOTA models to predict the same question. If the outputs of multiple SOTA models were consistent but differed from the reference answer, the reference answer was deemed incorrect, and the corresponding data was removed.
\end{itemize}

\subsection{Overlong Punishment}
 For the truncated samples, DAPO proposes two handling strategies: overlong filtering and soft overlong punishment. However, our experiments reveal that overlong filtering leads to a rapid increase in the output length, which is undesirable. Additionally, we observed that the performance of soft overlong punishment is comparable to directly treating overlong samples as incorrect answers. Therefore, in this work we simply treat overlong samples as incorrect answers and assigning them a reward of 0.

\section{Experiments}
\subsection{Training Details}
\label{sec:params}
In this section, we conducted experiments on Qwen3-30B-A3B to verify the effectiveness of UloRL. Experimental settings are represented as follows.

\textbf{Hyperparameter Settings}

For optimization, we utilize the AdamW optimizer~\citep{zhang2018international} with a constant learning rate of \(1 \times 10^{-6}\). During rollout, the prompt batch size is set to 128, and we sample 8 responses for each prompt. The sampling temperature is set to 0.85, with \(\text{top\_p}=1.0\) and \(\text{top\_k}=-1\). The maximum response length is set to 128k tokens, divided into a maximum of 8 segments, with each segment containing 16k tokens.

For training, the mini-batch size is set to 1024, meaning one gradient update is performed for each rollout step. The probability threshold for MPTs, \(\tau\), is set to 0.99, and the target entropy, \(\sigma\), is set to 0.2.

\textbf{Evaluation Setup}

For evaluation, we use the AIME-2025 and BeyondAIME\citep{yu2025dapo} datasets as benchmarks. Each evaluation set is repeated 32 times, and we report the average score (\(\text{avg@32}\)) to ensure result stability. The inference hyperparameters are set to a sampling temperature of 0.85, topp of 0.95 and topk of 20.

\subsection{Overall Results}
Table \ref{tab:overall} presents the evaluation results. The first group includes the performance metrics of SOTA models. The second group consists of three models tuned using different RL algorithms based on the Qwen3-30B-A3B model:
\begin{itemize}
    \item \textbf{UloRL-A3B-128k} This model is trained using the full UloRL algorithm, with training hyperparameters detailed in Section \ref{sec:params}.
    \item \textbf{UloRL-A3B-w/o-DMMPTs}: This is a variant of UloRL excluding the DMMPTs component. The training hyperparameters are identical to those used for the full UloRL method.
    \item \textbf{UloRL-A3B-128k-Yarn} Following \cite{Polaris2025}, we employ Yarn~\citep{peng2023yarn} to further extend the output length to 140k (factor=1.5,original\_len=93k).
\end{itemize}

\begin{table}[!t]
    \centering
        \begin{tabular}{l|c|c|c}
            \toprule
            \textbf{Model}  &\textbf{AIME-2025} & \textbf{BeyondAIME} & \textbf{AVG} \\
            \midrule
            DeepSeek-R1-0528 & 87.5 & 63.3$^*$ & 75.4\\
            Seed-1.6-thinking & 86 & 56.3 & 71.2 \\
            Qwen3-235B-A22B & 81.5 & 59.0$^*$ & 70.3 \\
            Qwen3-30B-A3B & 70.9 & 50.7$^*$ & 60.8 \\
            \hline
            \hline 
            UloRL-A3B-128k & 82.8 & 60.2 & 71.5 \\
            UloRL-A3B-w/o-DMMPTs & 78.6 & 57.1 & 67.9 \\
            UloRL-A3B-128k-Yarn & 85.1 & 61.9 & 73.5 \\
            \bottomrule
        \end{tabular}
    \caption{The overall results of the proposed UloRL trained on Qwen3-30B-A3B. Metrics marked with an * are results from our evaluation, while the others are from official reports.}
    \label{tab:overall}
\end{table}
From Table \ref{tab:overall}, we can make the following observations. (1) \textbf{UloRL-A3B-128k} outperforms \textbf{Qwen3-30B-A3B} with significant gains, even surpasses that of \textbf{Qwen3-235B-A22B}. These results confirm the effectiveness of the proposed \textbf{UloRL} algorithm , highlighting its ability to achieve state-of-the-art performance with a more efficient and scalable approach. (2) A comparison between \textbf{UloRL-A3B-w/o-DMMPTs} and \textbf{UloRL-A3B-128k} reveals that removing the DMMPTs strategy results in a significant degradation in model performance. This validates the efficacy of the proposed DMMPTs method. (3) By extending the output length to 140k using Yarn, the model achieved further improvements. This indicates that continuously expanding the length can further enhance the model's reasoning ability.

\subsection{Effect of Output Length on Model Performance}
\begin{table}[!t]
    \centering
        \begin{tabular}{l|c|c|c}
            \toprule
            \textbf{Model}  &\textbf{AIME-2025} & \textbf{BeyondAIME} & \textbf{AVG} \\
            \midrule
            Qwen3-30B-A3B & 70.9 & 50.7 & 60.8 \\
            \hline
            \hline
            UloRL-A3B-32k & 73.5 & 52.3 & 62.9 \\
            UloRL-A3B-64k & 79.9 & 58.5 & 69.2 \\
            UloRL-A3B-96k & 81.6 & 59.4 & 70.5\\
            \textbf{UloRL-A3B-128k} & \textbf{82.8} & \textbf{60.2} & \textbf{71.5} \\
            \bottomrule
        \end{tabular}
    \caption{The performances of models training with different output length.}
    \label{tab:length}
\end{table}

In this subsection, we investigate the effect of output length on model performance. We compared output lengths of 32k, 64k, 96k, and 128k. Except for the length and segment parameters, all other hyperparameters were kept consistent with those described in Section \ref{sec:params}. For the 32k experiment, the segment count was set to 1. For the 64k experiment, the output was divided into 4 segments, while for 96k and 128k, the output was divided into 8 segments.

The experimental results are shown in Table \ref{tab:length}. From the table, it can be observed that the performance improvement achieved with 32k reinforcement learning is minimal. This is primarily because Qwen3-30B-A3B is already a highly strong 32k-output model, and without significant changes to the output length, it is challenging to further enhance its reasoning capabilities. However, when the output length is extended to 64k, the model's reasoning ability improves significantly.

Overall, the results show a clear trend: the longer the output length, the better the model's reasoning performance. This demonstrates that extending the output length is an effective approach to improving the reasoning capabilities of large language models.

\section{Conclusions}
In this work, we proposed UloRL, an ultra-long output reinforcement learning algorithm for advancing Large Language Models' reasoning abilities. We first introduce the segment rollout to mitigate the inefficiencies caused by long-tail sequence distributions, enabling faster and more resource-efficient RL training. By incorporating Segment-Aware Importance Sampling (SAIS) and Pseudo On-Policy Importance Sampling (POIS), we ensure stable and accurate training dynamics in the segmented rollout setting. Furthermore, to tackle the issue of entropy collapse, we proposed the Dynamic Masking of well-Mastered Positive Tokens (DMMPTs) strategy, which adaptively balances exploration and exploitation without introducing additional optimization objectives or relying on importance sampling. Experimental results demonstrate the effectiveness of our methods. 

\clearpage
\bibliography{colm2024_conference}
\bibliographystyle{colm2024_conference}

\clearpage
\end{document}